\documentclass[11pt,a4paper]{article}
\usepackage[hyperref]{acl2021}
\usepackage{times}
\usepackage{latexsym}
\usepackage{amsmath}

\usepackage{microtype}

\aclfinalcopy 

\title{MIPT-NSU-UTMN at SemEval-2021 Task 5: Ensembling Learning with Pre-trained Language Models for Toxic Spans Detection}

\author{
  Mikhail Kotyushev \\
  Moscow Institute of \\Physics and Technology\\Moscow, Russia \\
  \texttt{mkotyushev@gmail.com} \\\And
  Anna Glazkova \\
  University of \\Tyumen\\Tyumen, Russia \\
  \texttt{a.v.glazkova@utmn.ru} \\\And
  Dmitry Morozov \\
  Novosibirsk State \\University\\Novosibirsk, Russia \\
  \texttt{morozowdm@gmail.com} \\}

\date{}

\begin{document}
\maketitle
\begin{abstract}
This paper describes our system for SemEval-2021 Task 5 on Toxic Spans Detection. We developed ensemble models using BERT-based neural architectures and post-processing to combine tokens into spans. We evaluated several pre-trained language models using various ensemble techniques for toxic span identification and achieved sizable improvements over our baseline fine-tuned BERT models. Finally, our system obtained a F1-score of 67.55\% on test data.
\end{abstract}

\section{Introduction}

Toxic speech has become a rising issue for social media communities. Abusive content is very diverse and therefore offensive language and toxic speech detection is not a trivial issue. Besides, social media moderation of lengthy comments and posts is often a time-consuming  process. In this regard, the task of detecting toxic spans in social media texts deserves close attention.

This work is based on the participation of our team, named MIPT-NSU-UTMN, in SemEval 2021 Task 5, ``Toxic Spans Detection'' \citep{pav2020semeval}. Organizers of the shared task provided participants with the trial, train, and test sets of English social media comments annotated at the span level indicating the presence or absence of text toxicity. We formulated the task as a token classification problem and investigated several BERT-based models using two-step knowledge transfer. We found that preliminary fine-tuning of the model on data that is close to the target domain improves the quality of the token classification. The source code of our models is available at \url{https://github.com/morozowdmitry/semeval21}.

The paper is organized as follows. A brief review of related work is given in Section 2. The definition of the task has been summarized in Section 3. The proposed methods and experimental settings have been elaborated in Section 4. Section 5 contains the results and error analysis respectively. Section 6 is a conclusion.

\section{Related Work}

Computational approaches to tackle text toxicity have recently gained a lot of interest due to the widespread use of social media. Since moderation is crucial to promoting healthy online discussions, research on toxicity detection has been attracting much attention. Our work is also related to hate speech and abusive language detection \citep{fortuna2020toxic}. 

The toxic speech detection task is usually framed as a supervised learning problem. Moreover, fairly generic features, such as bag of words \citep{harris1954distributional} or word embeddings \citep{mikolov2013efficient}, systematically yield reasonable classification performance \citep{fortuna2018survey,schmidt2017survey}. To better understand the mechanisms of toxic speech detection, some scholars \citep{waseem2017understanding,lee2018comparative,karan2018cross,swamy2019studying} compared different techniques for abusive language analysis. Neural architectures and deep learning methods achieved high results in this domain. Thus, \citet{pavlopoulos2017deep,pavlopoulos2017deeper} explored the possibilities of deep learning and deep attention mechanisms for abusive comment moderation. \citet{park2017one} proposed an approach to performing classification on abusive language based on convolutional neural networks (CNN). \citet{chakrabarty2019pay} used Bidirectional Long-Short Term
Memory network. \citet{castelle2018linguistic} experimented with CNN and Gated Recurrent Units. Some recent studies \citep{mozafari2019bert,risch2019hpidedis,liu2019nuli,nikolov2019nikolov} utilized pre-trained language models such as Bidirectional Encoder Representations from Transformers (BERT) \citep{DBLP:journals/corr/abs-1810-04805} to detect offensive or abusive language.

In recent years, the task of detecting and analyzing abusive, toxic, or offensive language has attracted the attention of more and more researchers. The shared tasks based on carefully curated resources, such as those organized at the SemEval \citep{zampieri2019semeval,basile2019semeval}, GermEval \citep{wiegand2018overview}, EVALITA \citep{bosco2018overview}, and OSACT \citep{mubarak2020overview} events, have significantly contributed to the progress of the field and to the enrichment of linguistic resources. In addition to the corpora collected for these shared tasks, \citet{rosenthal2020large} released a large-scale dataset for offensive language identification. \citet{ibrohim2018dataset,leite2020toxic,pitenis2020offensive,komalova2021mca} presented various datasets for abusive speech detection in non-English languages. Most of these datasets classify whole texts or documents, and do not identify the spans that make a text toxic.

\section{Shared Task}

The task focuses on the evaluation of systems that detect the spans that make a text toxic, when detecting such spans is possible. The goal of the task is to define a sequence of words (character offsets) that attribute to the toxicity of the text, for example:

\begin{itemize}
    \item \textbf{Input.} ``This is a \underline{stupid} example, so thank you for nothing \underline{a!@\#!@}''.
    \item \textbf{Output.} [10,11,12,13,14,15,51,52,53,54,55, 56].
\end{itemize}

The sources of data were various posts (comments) from publicly available datasets. The provided dataset contains 10,629 posts split into training (7939), trial (690), and test (2000) subsets.

Inspired by \citet{da2019fine}, the organizers proposed to employ the F1-score for evaluating the responses of a system participating in the shared task. Let system $A_i$ return a set $S^t_{A_i}$ of character offsets, for parts of the post found to be toxic. Let $G^t$ be the character offsets of the ground truth annotations of $t$. The F1-score of system $A_i$ is calculated with respect to the ground truth $G$ for post $t$ as follows, where $|\cdot|$ denotes set cardinality.

\begin{center}
$\textit{F}^t_1(A_i,G)=\frac{2\cdot P^t(A_i,G)\cdot R^t(A_i,G)}{P^t(A_i,G)+R^t(A_i,G)},$ \newline $\textit{P}^t(A_i,G)=\frac{S^t_{A_i}\cap S^t_G}{S^t_{A_i}},$\newline $\textit{R}^t(A_i,G)=\frac{S^t_{A_i}\cap S^t_G}{S^t_G}.$
\end{center}

The final F1-score is an average $\textit{F}^t_1(A_i,G)$ over all the posts $t$ of an evaluation dataset $T$ to obtain a single score for system $A_i$.

\section{Methodology}
The stated problem was modified from char-level to token-level binary-classification. The proposed solution utilizes a pre-trained language model with a classification head to classify tokens. Different configurations of BERT pre-trained as masked language models were considered as a backbone.

Due to the lack of available token-level labeled public datasets for toxic comment and the relatively small size and sparsity of dataset provided by the competition, the following training pipeline was proposed to enhance knowledge transfer. First, fine-tune pre-trained BERT on a larger-scale task of \textit{toxic comment classification}, using the Jigsaw dataset\footnote{\url{https://www.kaggle.com/c/jigsaw-toxic-comment-classification-challenge}} from which the competition data were constructed. Second, fine-tune obtained model to solve the actual \textit{toxic tokens classification} problem. The exact training parameters are to be found below.

For the first step:
\begin{itemize}
    \item remove texts occurred in spans dataset from classification dataset to prevent data leakage (so as spans dataset is sampled from classification dataset);
    \item 4 epochs, 200 tokens max length, 64 batch size, 10 gradient accumulation, mixed-precision FP16;
    \item default AdamW \citep{loshchilov2017decoupled} with lr = 4e-5, Layer-wise Decreasing Layer Rate \citep{DBLP:journals/corr/abs-1905-05583} with decay $\eta = 0.95$ and cosine learning rate (LR) schedule with T = 4 epochs and constant LR after epoch 3;
    \item selected bert-base-uncased as best performance / speed ratio;
    \item the best model on validation selection each 0.1 epoch by AUC.
\end{itemize}

For the second step:
\begin{itemize}
    \item hold-out $\approx14\%$ of data to train ensemble of models later;
    \item out-of-5-fold training on the residual $\approx86\%$ of data;
    \item 4 epochs, 512 tokens max length, 16 batch size, 10 gradient accumulation, mixed precision FP16;
    \item default AdamW with lr = 4e-5, Layer-wise Decreasing Layer Rate with decay $\eta = 0.95$ and cosine LR schedule with T = 4 epochs;
    \item the best model on validation selection each 0.1 epoch by F1-score.
\end{itemize}

The final solution contains $N \times K$ models, where $N$ is the number of different backbone BERT architectures, $K$ is the number of folds (5 in the current experiments). Obtained models are further to be ensembled using different strategies with validation on the single hold-out dataset:

\begin{itemize}
    \item hard voting: final spans are selected as at least one model (spans union), as all the models (spans intersection) or as some intermediate methods with at least $m$ models;
    \item soft voting: final probability is calculated as a weighted sum of models probabilities;
    \item train meta classifier.
\end{itemize}

\section{Experiments and Results}
Three pre-trained backbone BERT architectures were considered: BERT base uncased, BERT large uncased \citep{DBLP:journals/corr/abs-1810-04805}, and BERT base pre-trained for Hate Speech Detection \citep{aluru2020deep}. First step setup and results:
    
\begin{itemize}
    \item select subset of Jigsaw toxic classification data: all the targets with toxicity score $\geq$ 0.5 ($L = 135168$ objects) as class 1 and randomly sampled $3 * L$ objects with toxicity score $<$ 0.5 as class 0;
    \item stratified 80\% train, 20\% validation;
    \item 0.968 AUC bert-base, 0.968 AUC bert-large, 0.942 AUC dehate bert.
\end{itemize}

So as models except BERT base uncased did not show compatible performance for token classification (and later for tests on the fold 0 didn't show good F1-score for the actual task as well), later experiments were continued only for BERT base uncased pre-trained model fine-tuned on token classification.

For step two results are following:

\begin{itemize}
    \item train + trial, 8621 comments;
    \item average F1-score over 5 folds is 0.6714.
\end{itemize}

The experiments were conducted with Huggingface transformers library \citep{wolf2019huggingface}.

Many patterns in our results are expected, but some stand out. In general, our model is good at detecting obscene language and utterances that demean honor and dignity or denote low moral character. We noticed that our model is not very good at identifying the posts that have no toxic span annotations. According to the corpus description, in some toxic posts, the core message is conveyed may be inherently toxic. Thus, a sarcastic post can indirectly claim that people of a particular origin are inferior. Hence, it was difficult to attribute the toxicity of those posts to particular spans. In such cases, the corresponding posts were labeled as not containing toxic spans. Among our results, there are many examples where the model detected spans in not annotated posts, for example:

\begin{itemize}
\item ``uhhh Hillary Clinton is a serial killer and thief'': [] (true annotation), [26, 27, 28, 29, 30, 31, 33, 34, 35, 36, 37, 38, 44, 45, 46, 47, 48] (our annotation, ``uhhh Hillary Clinton is a \textbf{serial killer} and \textbf{thief}'');
\item ``This goes way beyond just being an asshole skipper, dude must have some serious mental issues'': [] (true annotation), [35, 36, 37, 38, 39, 40, 41] (our annotation, ``This goes way beyond just being an \textbf{asshole} skipper, dude must have some serious mental issues'').
\end{itemize}

In addition, some texts in the dataset raise questions of the annotation credibility, for example:

\begin{itemize}
    \item ``How the hell is this news?  Am I supposed to be shocked that the Crown Prince of Bahrain or one of the world's biggest celebrity superstars get's better access to the State \underline{Department} then I do?  During which administration has this ever not been true?  
    The media's desperation to keep this election close is far past ridiculous'' (training set, the toxic span annotation is underlined);
    \item ``Yup. NVN u\underline{sed the Pr}ess. The Press was USED. Used like their sister on prom night!
    \underline{Idiots}. All faux-erudite, not realizing they were being played'' (training set, the original annotation is underlined);
    \item ``And you are a complete \textbf{moron} w\underline{ho obv}iously doesn't know the meaning of the word \textbf{narcissist}. By the way your bias is showing'' (test set, the original annotation is underlined, the annotation of our model is highlighted in bold).
\end{itemize}

The final result of our model is presented in Table 1. As can be seen from the table, the systems of the participants produce close results. Our system achieved 67.55\% of F1-score on the test set of this shared task that attracted 91 submitted teams in total. This value exceeded the average result by almost 10\%. 

\begin{table}[]
\caption{Results on the test set.}
\begin{center}
\begin{tabular}{|l|l|l|}
\hline
Rank & Team & F1-score \\ \hline
1 & HITSZ-HLT & 0.7083 \\ \hline
26 & UAntwerp & 0.67552 \\ 
\textbf{27} & \textbf{MIPT-NSU-UTMN} & \textbf{0.67551} \\ 
28 & NLRG & 0.67532 \\ \hline
 & Avg result & 0.57805 \\ \hline
\end{tabular}
\end{center}
\end{table}

\section{Conclusion}

This paper introduces our BERT-based model for toxic spans detection. As expected, pre-training of the BERT model using an additional domain-specific dataset improves further toxic spans detection performance. Experimenting with different fine-tuning approaches has shown that our BERT-based model benefits from the two-step knowledge transfer technique. An ensemble with spans intersection obtained our best result on the test data.
 
In our future work, we will evaluate various language models, such as distilled versions of BERT \citep{sanh2019distilbert,jiao2020tinybert} and RoBERTa \citep{liu2019roberta}.


\bibliographystyle{acl_natbib}
\bibliography{anthology,acl2021}


\end{document}